%% file: Main.tex
\documentclass[onecolumn]{autart}

\usepackage[utf8]{inputenc} 
\usepackage[T1]{fontenc}    
\usepackage{hyperref}      
\usepackage{url}            
\usepackage{booktabs}       
\usepackage{amsfonts}       
\usepackage{nicefrac}       
\usepackage{array,multirow,graphicx}
\usepackage{float}
\usepackage{epsfig}
\usepackage{amsmath}
\usepackage{autobreak}
\usepackage{amssymb}
\usepackage{indentfirst}
\usepackage{mathrsfs}
\usepackage{cite}
\usepackage{bm}
\usepackage{graphicx}           
\usepackage{psfrag}             
\usepackage{subfigure}          
\usepackage{caption}
\usepackage{empheq}
\usepackage{enumerate}
\usepackage{soul}
\usepackage{bigints}

\usepackage[dvipsnames]{xcolor}
\usepackage{tikz}
\usetikzlibrary{arrows}

\newenvironment{proof}[1][Proof]%
  {\smallskip\par\noindent\textbf{#1\,:\ }}%
  {\hspace*{\fill} \rule{6pt}{6pt}\smallskip}
\newenvironment{proof*}[1][Proof]%
  {\medskip\par\noindent\textbf{#1\,:\ }}%

\newtheorem{assumption}{Assumption}
\newtheorem{condition}{Condition}
\newtheorem{theorem}{Theorem}
\newtheorem{lemma}{Lemma}

\usepackage{color,cite,verbatim}
\definecolor{gray}{RGB}{128,128,128}

\newcommand{\HB}[1]{{\color{orange} #1}}

\usepackage[utf8]{inputenc} 
\usepackage[T1]{fontenc}    
\usepackage{url}            
\usepackage{amsfonts}       
\usepackage{nicefrac}       
\usepackage{verbatim}

\usepackage{mathtools}
\DeclarePairedDelimiter{\ceil}{\lceil}{\rceil}

\usepackage{algorithm}

\usepackage{algorithmic}

\usepackage{setspace}
\let\Algorithm\algorithm
\renewcommand\algorithm[1][]{\Algorithm[#1]\setstretch{1}}

\usepackage{hyperref}

\begin{document}
\begin{frontmatter}
\title{Asynchronous Bayesian Learning over a Network}

\author[OSU]{K. Bhar}\ead{kbhar@okstate.edu},   
\author[OSU]{H. Bai}\ead{he.bai@okstate.edu},          \author[ARL]{J. George}\ead{jemin.george.civ@mail.mil},
\author[ARL]{C. Busart}\ead{carl.e.busart.civ@mail.mil}

\address[OSU]{Oklahoma State University, Stillwater, OK 74078, USA.}
\address[ARL]{U.S. Army Research Laboratory, Adelphi, MD 20783, USA.}             

\begin{keyword}                          
    Distributed Bayesian learning, Unadjusted Langevin algorithm, 
	Asynchronous Gossip protocol, Event-triggered mechanism, Multi-agent systems 
\end{keyword}                             

\begin{abstract}     
    We present a practical asynchronous data fusion model for networked agents to perform distributed Bayesian learning without sharing raw data.  Our algorithm uses a gossip-based approach where pairs of randomly selected agents employ unadjusted Langevin dynamics for parameter sampling.  We also introduce an event-triggered mechanism to further reduce communication between gossiping agents. These mechanisms drastically reduce communication overhead and help avoid bottlenecks commonly experienced with distributed algorithms. In addition, the algorithm’s reduced link utilization is expected to increase resiliency to occasional link failure.  We establish mathematical guarantees for our algorithm and demonstrate its effectiveness via numerical experiments.
\end{abstract}

\end{frontmatter}

\input{Intro}

\input{Prelims}
\input{Goss_ET}
\input{Results}

\input{Discussion}
\input{Num_Exp}
\input{Conclusion}
\newpage

\bibliographystyle{plain}
\bibliography{Ref}

\input{Appendix}

\end{document}

%% file: Intro.tex
\section{Introduction} \label{sec:Intro}

Distributed learning in machine learning applications have gained much attention recently due to ubiquitous applications in sensor networks and multi-agent systems. Often the data that a model needs to be trained on is distributed among multiple computing agents and it cannot be accrued in a single server location because of logistical constraints such as memory,  efficient data sharing means, or confidentiality requirements due to sensitive nature of the data. However, the need arises to train the same model with the entire distributed data. Isolated training individually by the agents with their local data may lead to overfitted models as the training data is limited. {Besides, training such isolated models on different agents is redundant as more parameter updates have to be performed by the isolated models to reach a certain level of accuracy as compared to what can be achieved by sharing information.}
Distributed learning aims to leverage the full distributed data by a coordinated training among all the agents where the agents are allowed to share partial information (usually the learned model parameters or their gradients) without sharing any raw data. The information shared is significantly lower compared to sharing the raw data and does not compromise  confidentiality.

In this paper, we focus on Bayesian inference techniques. Bayesian learning has been established as a reliable method for training machine learning models involving large datasets and a large number of trainable parameters. {Additionally, since Bayesian inference techniques are based on \emph{sampling} from posterior probabilities, they provide a built-in mechanism to quantify uncertainty.} Because computing exact posteriors in most practical scenarios is analytically or computationally impossible, {one of the common industry standards is to implement Markov Chain Monte Carlo (MCMC) sampling methods}. In this paper, we employ the unadjusted Langevin algorithm (ULA) as the sampling method. Centralized Langevin methods are well studied. Convergence of such algorithms for strongly log-concave target distributions~\cite{dalalyan2017theoretical, dalalyan2017further, cheng2018underdamped, cheng2018convergence, durmus2016sampling, durmus2017nonasymptotic, durmus2019high} as well as for non-convex cases like in~\cite{raginsky2017non, xu2018global, zhang2019nonasymptotic, chau2021stochastic, mou2019improved, majka2020nonasymptotic, ma2019sampling, cheng2018sharp} has been established.

Unsurprisingly, distributed~\cite{parayil2020decentralized, kungurtsev2021decentralized, kolesov2021decentralized,gurbuzbalaban2021decentralized} and federated~\cite{lee2020bayesian, el2021federated} formulations of various Bayesian based algorithms have been developed as well. However, most literature on distributed Bayesian learning deals with synchronized updates by all agents at any given time~\cite{parayil2020decentralized, kungurtsev2021decentralized, kolesov2021decentralized,gurbuzbalaban2021decentralized}, which is not practical in real scenarios as the algorithm is expected to suffer from link failures. The synchronized updates may be stymied due to lagging agents while faster agents sit idle. Also since the synchronized update by any agent depends on the shared information from its neighbors, there is immense communication overhead at every time instant. 
We seek to develop an algorithm to circumvent the aforementioned shortcomings. Drawing inspiration from traditional optimization literature~\cite{boyd2005gossip, ram2009asynchronous}, we introduce the concept of asynchronous gossip updates to the Bayesian ULA. The gossip algorithm allows asynchronous updates where at any time a single agent is randomly active, it randomly chooses one of its neighbors to share information, and together they make a single update step. Thus, at any particular time only two agents are active. An inherent assumption of gossip algorithms is that no two agents become active at the same time. Since at most a single link is active at any time, it is more robust to occasional link failures mentioned earlier. Also, communication overhead is drastically reduced. 

Furthermore, we incorporate an event-triggered  information sharing scheme where information  between the two active agents does not need to be transferred unless some event is triggered. This further mitigates the communication overhead issue. We present rigorous convergence proofs for the proposed algorithm. The results obtained in this paper are of  practical relevance as they model the information exchange over a graph much more pragmatically. To make the updates truly asynchronous, we propose using a constant step size. Using constant step sizes results in a bias in the convergence, which has been shown to exist even for centralized implementations~\cite{wibisono2018sampling}. We discuss in details guidelines to minimize the bias in the convergence. We also support our results by providing {two} illustrative examples.

The rest of the paper is organized as follows. We start with an introduction of the Bayesian learning framework and the ULA utilized for Bayesian inference in Section~\ref{sec:Prelims}. In Section~\ref{sec:Goss_ET}, we introduce the key aspects of the gossip protocol and the event-triggering scheme. In Section~\ref{sec:Results}, we state our mathematical guarantees. Section~\ref{sec:Discussions} provides further insight of our results, while two numerical simulation examples are illustrated in Section~\ref{sec:Num_Exp}. Finally, we conclude with Section~\ref{sec:Conclusion}.

\textbf{Notation}:  Let $\mathbb{R}^{n\times m}$ denote the set of $n\times m$ real matrices. For a vector $\bm{\phi}$, $\phi_i$ is the ${i^{th}}$ entry of $\bm{\phi}$. An $n\times n$ identity matrix is denoted as $I_n$. $\mathbf{1}_n$ denotes a $n$-dimensional vector of all ones and $\mathbf{e}_i$ is a $n$-dimensional vector with all $0$s except the $i$-th element being $1$. The $p$-norm of a vector $\mathbf{x}$ is denoted as $\left\| \mathbf{x} \right\|_p$ for $p \in [1,\infty]$. Given matrices $A \in \mathbb{R}^{m\times n}$ and $B \in \mathbb{R}^{p \times q}$, $A \otimes B \in \mathbb{R}^{mp \times nq}$ denotes their Kronecker product. For a graph $\mathcal{G}\left(\mathcal{V},\mathcal{E}\right)$ of order $n$, $\mathcal{V} \triangleq \left\{v_1, \ldots, v_n\right\}$ represents the agents or nodes and the communication links between the agents are represented as $\mathcal{E} \triangleq \left\{\varepsilon_1, \ldots, \varepsilon_{\ell}\right\} \subseteq \mathcal{V} \times \mathcal{V}$. Let $\mathcal{A} = \left[a_{i,j}\right]\in \mathbb{R}^{n\times n}$ be the \emph{adjacency matrix} with entries of $a_{i,j} = 1 $ if $(v_i,v_j)\in\mathcal{E}$ and zero otherwise. Define $\Delta = \text{diag}\left(\mathcal{A}\mathbf{1}_n\right)$ as the in-degree matrix and $\mathcal{L} = \Delta - \mathcal{A}$ as the graph \emph{Laplacian}. A Gaussian distribution with a mean $\mu\in\mathbb{R}^m$ and a covariance $\Sigma\in\mathbb{R}_{\geq0}^{m\times m}$ is denoted by $\mathcal{N}(\mu,\Sigma)$.

%% file: Prelims.tex
\section{Preliminaries} \label{sec:Prelims}

\subsection{Bayesian inference framework}
Consider a network of $n$ agents characterized by an undirected communication graph $\mathcal{G}(\mathcal{V},\mathcal{E})$ of order $n$. The entire data $\bm{X} = \{\bm{X}_i\}_{i=1}^{n}$ is distributed among $n$ agents with the $i$-th agent having access only to its local dataset $\bm{X}_i=\{x_i^j\}_{j=1}^{M_i}$, where $x_i^j \in \mathbb{R}^d$. Since individual agents do not have access to others' datasets, proper fusion and update  to distributedly infer common parameters is a non-trivial task of paramount practical significance.

Bayesian learning provides a framework for leaning unknown parameters by sampling from a posterior distribution. The probability of the unknown parameter $\bm{w}$ given the data $\bm{X}$, denoted by $p(\bm{w}|\bm{X})$, is the posterior distribution of interest. Assuming that the individual datasets of the agents are conditionally independent, the target posterior distribution $p^*(\bm{w})\triangleq p(\bm{w}|\bm{X})$ is given by 
\begin{align}
    &p(\bm{w}|\bm{X}) \propto p(\bm{w}) \prod_{i=1}^{n} p(\bm{X}_i|\bm{w}) = \prod_{i=1}^{n} p(\bm{X}_i|\bm{w}) p(\bm{w})^{\frac{1}{n}}. \label{eq:global_posterior}
\end{align}
Thus, the objective of the inference problem is to determine $p^*$. As analytical solutions to $p^*$ are often intractable, MCMC algorithms aim at sampling from $p^*$.  

\subsection{Sampling method}
We use the unadjusted Langevin algorithm (ULA) which is a first order gradient method for sampling from $p^*$. Define an energy function $E(\bm{w}) = -\log(p(\bm{w}|\bm{X}))$. It follows from~\eqref{eq:global_posterior} that for some constant $C$,
\begin{align}
    E(\bm{w},\bm{X}) &= \sum_{i=1}^{n} E_i(\bm{w},\bm{X}_i) + C, \label{eq:energy_function}
\end{align}
where $E_i(\bm{w}) = - \log p(\bm{X}_i|\bm{w}) - \frac{1}{n} \log p(\bm{w})$.
 In the centralized sampling scenario, the ULA can be represented as 
\begin{align}
    &\bm{w}(k+1) = \bm{w}(k) - \alpha \nabla E(\bm{w}(k),\bm{X}) + \sqrt{2\alpha} \bm{v}(k), \label{eq:ULA}
\end{align}
where $\alpha>0$ 
is the gradient step size, the gradient is given as $\nabla E = -\nabla \log p\left(\bm{X}| \bm{w}\right) - \nabla \log p(\bm{w})$, and $\bm{v}(k) \sim \mathcal{N}(\mathbf{0}_{d_w},I_{d_w})$ is an injected Gaussian noise. A distributed version of~\eqref{eq:ULA} was introduced in~\cite{parayil2020decentralized} which is given by
\begin{align}
\begin{split}
    &\bm{w}_i(k+1) = \bm{w}_i(k) - \beta_k \sum_{j\in \mathcal{N}_i} \left(\bm{w}_i(k) - \bm{w}_j(k)\right) - \alpha_k n \nabla E_i(\bm{w}_i(k),\bm{X}_i) + \sqrt{2 \alpha_k} \bm{v}_i(k), \label{eq:DULA}
\end{split}
\end{align}
where $\bm{w}_i(k)$ is the sample of the $i$-th agent, $\mathcal{N}_i$ denotes the set of neighbors of the $i$-th agent, $\alpha_k$ 
is the time-dependent gradient step size, $\beta_k$ 
is a time-dependent fusion weight, the individual agent's gradients are given as $\nabla E_i = -\nabla \log p\left(\bm{X}_i| \bm{w}_i\right) - \frac{1}{n} \nabla \log p(\bm{w}_i)$, and $\bm{v}_i(k) \sim \mathcal{N}(\mathbf{0}_{d_w},nI_{d_w})$.

%% file: Goss_ET.tex
\section{Asynchronous Gossip with Event-triggering} \label{sec:Goss_ET}

\subsection{Gossip protocol}
One of the major drawbacks of the algorithm in~\eqref{eq:DULA} is the communication overhead presented by the fusion term $\sum_{j \in \mathcal{N}_i}\left(\bm{w}_i(k) - \bm{w}_j(k)\right)$. This necessitates communication between all the neighbors at all time instants in a synchronized fashion. Motivated by the optimization literature, we introduce the asynchronous gossip protocol~\cite{ram2009asynchronous} which circumvents this issue by needing only $2$ agents to update their samples at any given time instant.

Consider that each agent has local clock that ticks at a Poisson rate of $1$. At each tick of its clock, it randomly chooses one of its neighbors and together they make updates. {We assume that no two ticks of the local clocks of the agents coincide.} For the purpose of analysis, we consider a universal clock which ticks at a rate of $n$ and is indexed by $k$. Suppose that the $k$-th tick of the universal clock coincides with the $i_k$-th agent's local clock. Then agent $i_k$ chooses agent $j_k$ from $\mathcal{N}_{i_k}$ uniformly at random to communicate. The probability of agent $i$, $\forall\, i\in\{1,\cdots,n\}$, being active at the $k$-th tick of the universal clock is given by $p_i = \frac{1}{n} \left(1 + \sum_{j\in \mathcal{N}_i} \frac{1}{|\mathcal{N}_j|} \right)$. Note that $p_i$, $\forall i$, is time-invariant and depends on the graph only. Thus, it can be computed and stored by each agent a priori and subsequently used when needed.

Let $\mathcal{A}_k = \{i_k,j_k\}$ be the set of two agents activated at the $k$-th tick of the universal clock. Denote by $\tau_i(k)$ the number of times agent $i$ has been active until the $k$-th tick of the universal clock. The update algorithm for the active agents, i.e., $i \in \mathcal{A}_k$, is given by 
\begin{align}
    &\bm{w}_i(\tau_i(k)+1) = \bm{w}_i(\tau_i(k)) - \beta \sum_{j\in\mathcal{A}_k} (\bm{w}_i(\tau_i(k)) - \bm{w}_j(\tau_j(k))) - \frac{n\alpha}{2p_i} \nabla E_i(\bm{w}{_i}(\tau_i(k)),\bm{X}_i) + \sqrt{2\alpha} \bm{v}_i(\tau_i(k)), \label{eq:DULA_goss_inv}
\end{align}
where  $\alpha$ and $\beta$ are constant gradient step size and fusion weight, respectively, $\nabla E_i = - \nabla\log p\left(\bm{X}_i| \bm{w}_i\right) - \frac{1}{n} \nabla \log p(\bm{w}_i)$, and $\bm{v}_i$ is the injected noise given by $\bm{v}_i \sim \mathcal{N}\left( \mathbf{0}_{d_w}, \frac{n^2}{2} I_{d_w} \right)$. Define $\delta_i(k)$ as the indicator function such that $\delta_i(k) = 1$ if $i \in \mathcal{A}_k$ and otherwise $\delta_i(k) = 0$. Thus, for agent $i$, $\forall\, i\in \{1, \cdots, n\}$, the gossip-based sampling protocol~\eqref{eq:DULA_goss_inv} can be represented in the universal clock index $k$ as
\begin{align}
\begin{split}
    &\bm{w}_i(k+1) = \bm{w}_i(k) - \delta_i(k) \beta \sum_{j\in \mathcal{A}_k} \left(\bm{w}_i(k) - \bm{w}_j(k)\right) - \delta_i(k) \frac{n\alpha}{2p_i} \nabla E_i(\bm{w}{_i}(k), \bm{X}_i) + \delta_i(k) \sqrt{2\alpha} \bm{v}_i(k). \label{eq:DULA_gossip}
\end{split}
\end{align}
{For any agent $i$, $\bm{w}_i(\tau_i(k)) = \bm{w}_i(k)$. For all the ticks of the universal clock between the $\tau_i(k)^{\text{th}}$ and the $(\tau_i(k)+1)^{\text{th}}$ ticks of the $i$-th agent's local clock, $\bm{w}_i(k)$ remains unchanged. Note that we represent the algorithm in the universal clock index in~\eqref{eq:DULA_gossip} only for the purpose of our analysis. The individual agents do not need the ticks of the universal clock.}

\subsection{Event-triggering mechanism}
The gossip protocol in~\eqref{eq:DULA_gossip} allows asynchronous updates between agents and drastically reduces communication overhead. We next introduce an event-triggering mechanism that further reduces the need to exchange samples at all the time instants each agent is active. Unless an agent is triggered, it does not communicate its sample to its gossiping neighbor and the neighbor proceeds with the last communicated sample of that agent. Denote by $\hat{\bm{w}}_i(k)$ the last communicated sample of $i$-th agent until the the $k$th tick of the universal clock. Agent $i$ is triggered again to communicate $\bm{w}_i(k)$ if and only if $\delta_{i}(k)=1$ and
\begin{align}
    \|\bm{e}_i(k)\|_2^2 = \|\bm{w}_i(k) - \hat{\bm{w}}_i(k)\|_2^2 > \epsilon_i(k). \label{eq:ET}
\end{align}
Incorporating the event-triggering mechanism~\eqref{eq:ET} into~\eqref{eq:DULA_gossip}, we propose the following sampling algorithm for agent $i$, $\forall i$
\begin{align}
\begin{split}
    &\bm{w}_i(k+1) = \bm{w}_i(k) - \delta_i(k) \beta \sum_{j\in \mathcal{A}_k} \left(\hat{\bm{w}}_i(k) - \hat{\bm{w}}_j(k)\right) - \delta_i(k) \frac{n\alpha}{2p_i} \nabla E_i(\bm{w}{_i}(k),\bm{X}_i) + \delta_i(k) \sqrt{2\alpha} \bm{v}_i(k). \label{eq:DULA_gossip_ET}
\end{split}
\end{align}
We choose the triggering threshold $\epsilon_i(k)$ as
\begin{align}
    &\epsilon_i(k) = \frac{\mu^e_i}{(\tau_i(k)+1)^{\delta_i^e}} \leq \frac{\mu_e}{(k+1)^{\delta_e}}, \label{eq:ET_bound}
\end{align}
{where $\delta_i^e>0$ and $\mu^e_i>0$ are agent-specific parameters independently chosen to control the event-triggering rate, while $\mu_e = \frac{1}{(2n)^{\delta_e}}  \max_{i}\{\mu^e_i\} >0$ and $\delta_e = \min_{i} \{\delta_i^e\}>0$. The last inequality in~\eqref{eq:ET_bound} holds for sufficiently large $k$ with probability $1$ (see~\cite[Lemma 3]{nedic2010asynchronous}).}

%% file: Results.tex
\section{Results} \label{sec:Results}

We present the key results of our analysis in this section. In Section~\ref{sec:con_avg_dyn}, we derive from~\eqref{eq:DULA_gossip_ET} the consensus dynamics and the average dynamics to lay the foundation of the consensus and convergence analysis, respectively.  In Section~\ref{sec:Result_assumption}, we state the assumptions and the conditions needed for our analysis.  Section~\ref{sec:Result_consensus} and~\ref{sec:Result_convergence} present the main results pertaining to the consensus and the convergence, respectively.

\subsection{Consensus and average dynamics} \label{sec:con_avg_dyn}
We define the following notations: $\mathbf{w}(k) = \left[\bm{w}_1(k)^\top,\ldots,\bm{w}_n(k)^\top \right]^\top$, $\mathbf{v}(k)= \big[\bm{v}_1(k)^\top,\ldots, \bm{v}_n(k)^\top \big]^\top$, $\mathbf{e}(k) = \big[\bm{e}_1(k)^\top,$ \\ $\ldots,\bm{e}_n(k)^\top \big]^\top$ and $\nabla \mathbf{E}(k)= \left[ \nabla E_1(\bm{w}_1(k),\bm{X}_1){^\top}, \ldots, \nabla E_n(\bm{w}_n(k),\bm{X}_n){^\top} \right]^\top.$
We rewrite~\eqref{eq:DULA_gossip_ET} in the vector form as 
\begin{align}
\begin{split}
    \mathbf{w}(k+1) = \mathcal{W}_{{k}} \mathbf{w}{(k)} - \alpha n S_k \nabla \mathbf{E}(k) + \sqrt{2\alpha} {S'_k} \mathbf{v}(k) + \beta {(\mathcal{L}_k\otimes I_{d_w})}\mathbf{e}(k), \label{eq:Algo_compact}
\end{split}
\end{align}
where $\mathcal{L}_k= (\mathbf{e}_{i_k}- \mathbf{e}_{j_k}) (\mathbf{e}_{i_k}-\mathbf{e}_{j_k})^{\top}$, $\mathcal{W}_k= \big(I_n- \beta \mathcal{L}_k\big)\otimes I_{d_w} $, $S_k = \left( {\frac{1}{2p_{i_k}}} \mathbf{e}_{i_k} \mathbf{e}_{i_k}^{\top} + {\frac{1}{2 p_{j_k}}} \mathbf{e}_{j_k}\mathbf{e}_{{j}_k}^{\top} \right)\otimes I_{d_w}$ and $S'_k = (\mathbf{e}_{i_k} \mathbf{e}_{i_k}^{\top} + \mathbf{e}_{j_k}\mathbf{e}_{{j}_k}^{\top})\otimes I_{d_w}$. Let $\bar{\bm{w}}(k) = \frac{1}{n} \sum_{i=1}^{n} \bm{w}_i(k)$. We define the consensus error $\tilde{\mathbf{w}}(k) = \big[ (\bm{w}_1(k)-\bar{\bm{w}}(k))^\top, \ldots, (\bm{w}_n(k)- \bar{\bm{w}}(k))^\top \big]^\top$ and note that 
\begin{align}
    \tilde{\mathbf{w}}(k) = (M \otimes I_{d_w}) \mathbf{w}(k),\label{eq:con_err_def}
\end{align}
where $M = I_n - \frac{1}{n}\mathbf{1}_n \mathbf{1}_n^\top$.
Pre-multiplying~\eqref{eq:Algo_compact} with $(M \otimes I_{d_w})$ yields the evolution of the consensus dynamics:
\begin{align}
    \tilde{\mathbf{w}}(k+1) = \mathcal{W}_{{k}} \tilde{\mathbf{w}}(k) + (M\otimes I_{d_w})\mathbf{g}(k), \label{eq:consensus_err}
\end{align}
where $\mathbf{g}(k) = -\alpha n S_k \nabla\mathbf{E}(k) + \sqrt{2\alpha} {S'_k} \mathbf{v}(k) + \beta {(\mathcal{L}_k \otimes I_{d_w})} \mathbf{e}(k)$ and $(M \otimes I_{d_w}) \mathcal{W}_{{k}} = \mathcal{W}_{{k}} (M \otimes I_{d_w})$. 

Next, we derive the dynamics of the averaged sample $\bar{\bm{w}}(k)$ generated at each tick of the universal clock as
\begin{align}
    &\bar{\bm{w}}(k+1) = \bar{\bm{w}}(k) - \alpha \widehat{\nabla E}(k) + \sqrt{2\alpha} \bar{\bm{v}}(k), \label{eq:average_dynamics}
\end{align}
where $\widehat{\nabla E}(k) = \sum_{i\in \mathcal{A}_k} \nabla E_i(\bm{w}{_i}(k),\bm{X}_i)$ and $\bar{\bm{v}}(k) = \frac{1}{n} \sum_{i\in \mathcal{A}_k} \bm{v}_i(k) \sim \mathcal{N}\left(\mathbf{0}_{d_w}, I_{d_w} \right)$. The $\widehat{\nabla E}(k)$ can be considered a stochastic gradient and is related to the full gradient 
$\nabla E(\bar{\bm{w}}(k)) = \sum_{i=1}^n \nabla E_i(\bar{\bm{w}}(k),\bm{X})$ by \begin{align}
    &\widehat{\nabla E}(k) = \nabla E(\bar{\bm{w}}(k)) - \xi( \bar{\bm{w}}(k),\mathcal{A}_k) + \zeta(\bar{\bm{w}}(k), \tilde{\mathbf{w}}(k), \mathcal{A}_k), \label{eq:grad_split}
\end{align}
where 
\begin{align}
    \xi(\bar{\bm{w}}(k),\mathcal{A}_k) &= \nabla E(\bar{\bm{w}}(k)) - \sum_{i\in\mathcal{A}_k} \frac{1}{2p_i} \nabla E_i(\bar{\bm{w}}(k),\bm{X}_i), \label{eq:xi_def} 
\end{align}
\begin{align}
    \begin{split}
    \zeta(\bar{\bm{w}}(k), \tilde{\mathbf{w}}(k), \mathcal{A}_k) &= \sum_{i\in\mathcal{A}_k} \frac{1}{2p_i} \bigg( \nabla E_i(\bm{w}_i(k),\bm{X}_i)- \nabla E_i(\bar{\bm{w}}(k), \bm{X}_i) \bigg). \label{eq:zeta_def}
    \end{split}
\end{align}
The $\xi(k)$ represents the stochasticity from the gossip protocol while $\zeta(k)$ denotes the gradient noise due to consensus error. It follows that $\mathbb{E}_{p_t(\mathcal{A}_k)}[\xi(k)] = 0$.

\subsection{Assumptions and Conditions} \label{sec:Result_assumption}
Below we state all the assumptions needed and the conditions on the parameters that are essential to conclude the convergence results.

\begin{assumption} \label{assump:Lips_gradient}
The gradients $\nabla E_i$ are Lipschitz continuous with Lipschitz constant $L_i>0$ for all $i \in \{1,\ldots,n\}$, i.e., $\forall \, \bm{w}_a, \bm{w}_b \in \mathbb{R}^{d_w}$, we have 
\begin{align}
    \|\nabla E_i(\bm{w}_a, \bm{X}_i) - \nabla E_i(\bm{w}_b, \bm{X}_i) \|_2 \leq L_i \| \bm{w}_a - \bm{w}_b\|_2. \label{eq:Lip_assump}
\end{align}
\end{assumption}
From~\eqref{eq:Lip_assump} it follows that for $E(\bm{w}, \bm{X})$ in~\eqref{eq:energy_function}, there exists some $\bar{L}>0$ such that $\forall \, \bm{w}_a, \bm{w}_b \in \mathbb{R}^{d_w}$ we have 
\begin{align}
    \|\nabla E(\bm{w}_a,\bm{X}) - \nabla E(\bm{w}_b,\bm{X})\|_2 \leq \bar{L} \|\bm{w}_a - \bm{w}_b\|_2.
\end{align}
For the function $G(\mathbf{w},\bm{X})$ defined as 
\begin{align}
    G(\mathbf{w},\bm{X}) = \sum_{i=1}^{n} \nabla E_i(\bm{w}_i, \bm{X_i}),
\end{align}
where $\mathbf{w} = \big[\bm{w}_1^\top, \ldots, \bm{w}_n^\top \big]^\top \in \mathbb{R}^{nd_w}$, we also conclude from~\eqref{eq:Lip_assump} that there exists $L = \max_{i} \{L_i\} >0$ such that $\forall \, \mathbf{w}_a, \mathbf{w}_b \in \mathbb{R}^{nd_w}$ we have 
\begin{align}
    \|G(\mathbf{w}_a,\bm{X}) - G(\mathbf{w}_b,\bm{X})\|_2 \leq L \|\mathbf{w}_a - \mathbf{w}_b \|_2.
\end{align}

\begin{assumption} \label{assump:graph}
The overall interaction topology of the $n$ networked agents is given as a \emph{connected}, \emph{undirected} graph denoted by $\mathcal{G}(\mathcal{V}, \mathcal{E})$.
\end{assumption}
For a connected undirected graph $\mathcal{G}(\mathcal{V}, \mathcal{E})$,  the expected graph Laplacian, denoted by  $\bar{\mathcal{L}} = \mathbb{E}[\mathcal{L}_k]$, is a positive semi-definite matrix with exactly one eigenvalue at $0$ corresponding to the eigenvector $\mathbf{1}_n$.

\begin{assumption} \label{assump:bounded_grad}
There exists some $0<\mu_g<\infty$ such that for any $\bm{w} \in \mathbb{R}^{d_w}$, we have
\begin{align}
    \sup_{i\in\{1,\ldots, n\}} \mathbb{E}[\|\nabla E_i(\bm{w},\bm{X})\|_2] \leq \sqrt{\mu_g}.
    \label{eq:bounded_grad}
\end{align}
\end{assumption}
Also, \eqref{eq:bounded_grad} can be equivalently represented as
\begin{align}
    \mathbb{E} [\|\nabla \mathbf{E}(\mathbf{w}, \bm{X})\|_2^2] \leq n\mu_g.
\end{align}
Note that Assumption~\ref{assump:bounded_grad} is a standard assumption in many optimization literature.

\begin{assumption} \label{assump:bounded_moment_sto}
We assume that the second moment of the stochastic noise due to gossip in the average gradient $\xi$ is bounded, i.e., for all $k\geq0$ there exists some $0<C_{_{\xi}}<\infty$ such that 
\begin{align}
    \mathbb{E}_{p_t(\mathcal{A}_k)}[\| \xi(\bar{\bm{w}}(k), \mathcal{A}_k)\|_2^2] \leq C_{_{\xi}}.
\end{align}
\end{assumption}

\begin{assumption} \label{Assump:LSI}
The target distribution $p^*$  satisfies a \emph{$\log$-Sobolev inequality} (LSI) defined as follows. For any smooth function $g$ satisfying $\int \, g(\bar{\bm{w}}) p^*(\bar{\bm{w}}) \, d\bar{\bm{w}} = 1$, a constant $\rho_U > 0$ exists such that 
\begin{equation} 
\begin{split}
    \int g(\bar{\bm{w}}) \log g(\bar{\bm{w}}) & {p}^{*}(\bar{\bm{w}})\, d\bar{\bm{w}} \leq \frac{1}{2\rho_U}\int \frac{\left\| \nabla g (\bar{\bm{w}})\right\|^2}{g(\bar{\bm{w}})} {p}^{*}(\bar{\bm{w}})\, d\bar{\bm{w}},     
\end{split}\label{eq:LSI_original}
\end{equation}
where $\rho_U$ is the log-Sobolev constant.
\end{assumption}

\begin{condition} \label{cond:alpha_cond}
{The step size $\alpha$ is chosen to satisfy} 
\begin{align}
    \frac{8\alpha^3\bar{L}^4}{(1- \exp(-\alpha\rho_U))} < \rho_U. \label{eq:alpha_cond}
\end{align}
\end{condition}

\begin{condition} \label{cond:beta_cond}
{The fusion weight $\beta$ is chosen to satisfy}
\begin{align}
    \beta(1-\beta) < \frac{1}{2\lambda_{n-1}(\bar{\mathcal{L}})}, \label{eq:beta_cond}
\end{align}
where $\lambda_{n-1}(\bar{\mathcal{L}})$ is the second smallest eigenvalue of $\bar{\mathcal{L}}$.
\end{condition}

Note that the left hand side of~\eqref{eq:alpha_cond} decreases monotonically with a decreasing $\alpha$ and approaches $0$ as $\alpha$ approaches $0$. Thus, given a $\rho_U$, there always exists an $\alpha^*>0$ such that for any $\alpha\in(0,\alpha^*]$,~\eqref{eq:alpha_cond} holds. Similarly, for a given graph, $\frac{1}{2\lambda_{n-1}(\bar{\mathcal{L}})}$ is constant while $\beta(1-\beta)$ approaches $0$ as $\beta$ approaches $0$. Thus, there always exists a $\beta^*>0$ such that~\eqref{eq:beta_cond} holds for any $\beta\in(0,\beta^*]$. 

\subsection{Consensus analysis} \label{sec:Result_consensus}
Theorem~\ref{thm:consensus} below shows that consensus is achieved at the rate of $\mathcal{O} \left( \frac{1}{k^{\delta_e}}\right)$ with an offset $Y_3$ given in~\eqref{eq:O_def}. We refer to Section~\ref{sec:Discussions} for further discussions on the convergence.

\begin{theorem}\label{thm:consensus}
Suppose that Assumptions~\ref{assump:Lips_gradient}--\ref{assump:bounded_moment_sto} hold and that $\alpha$ and $\beta$ satisfy Conditions~\ref{cond:alpha_cond} and~\ref{cond:beta_cond}, respectively. Define $\lambda = 1 - 2\beta(1 - \beta) \lambda_{n-1}(\bar{\mathcal{L}})$ where $\lambda_i(\cdot)$ denotes the $i$-th largest eigenvalue of the positive semi-definite matrix $\bar{\mathcal{L}} = \mathbb{E}[\mathcal{L}_k]$. Then the consensus error $\tilde{\mathbf{w}}(k+1)$ defined in~\eqref{eq:con_err_def} satisfies
\begin{align}
    &\mathbb{E}[\|\tilde{\mathbf{w}}(k+1)\|_2^2] \leq Y_1 \sqrt{\lambda}^{k+1} + \frac{Y_2}{(k+1)^{\delta_e}} + Y_3,
    \label{eq:consensus_error_bound}
\end{align}
where the positive constants $Y_1$, $Y_2$, and $Y_3$ are given by
\begin{align}
    Y_1 &= \mathbb{E}[\|\tilde{\mathbf{w}}(0)\|_2^2] + \frac{2{\beta}^2n\mu_e}{1-\sqrt{\lambda}} \sum_{t=0}^{\bar{t}-1} \frac{\sqrt{\lambda}^{-(t+1)}}{(t+1)^{\delta_e}}, \label{eq:Y1_def} \\
    \quad\bar{t} &= \max\Bigg\{0, \ceil[\Big]{\frac{\delta_e}{|\ln{\lambda}|}-1} \Bigg\}, \label{eq:tbar_def}\\
    Y_2 &= - \frac{2{\beta}^2n\mu_e}{\sqrt{\lambda}(1-\sqrt{\lambda})} \left(\ln{\sqrt{\lambda}} + \frac{\delta_e}{\bar{t}+ 1}\right)^{-1}, \label{eq:Y2_def} \\
    Y_3 &= \frac{2\alpha n^2(\frac{\alpha\mu_g}{4p_m^2}+2d_w)}{(1 - \sqrt{\lambda})^2}, \label{eq:O_def} 
\end{align}
where $p_m=\min_{i} p_i$.
\end{theorem}
\begin{proof}
To analyze the consensus error, we start with the consensus dynamics in~\eqref{eq:consensus_err} and take the $L_2$ norm on both sides, yielding
\begin{align}
    \|\tilde{\mathbf{w}}(k+1)\|_2 \leq \|\mathcal{W}_{{k}} \tilde{\mathbf{w}}(k)\|_2 + \|\mathbf{g}(k)\|_2, \label{eq:consensus_err_norm}
\end{align}
where we used the result $\|(M \otimes I_{d_w}) \mathbf{g}(k)\|_2 \leq \|M \otimes I_{d_w}\|_2 \|\mathbf{g}(k)\|_2 = \|\mathbf{g}(k)\|_2$ since $\|M \otimes I_{d_w}\|_2=1$. Denoting by $\mathcal{F}_k$ be the filtration generated by randomized sampling of $\{\mathbf{w}(\ell)\}_{\ell=0}^{k}$, it can be shown that the conditional expectation $\mathbb{E}[\|\mathcal{W}_k \tilde{\mathbf{w}}(k)\|_2^2| \mathcal{F}_k]$ follows the relation below: 
\begin{align}
    &\mathbb{E}[\|\mathcal{W}_k \tilde{\mathbf{w}}(k)\|_2^2| \mathcal{F}_k] = \tilde{\mathbf{w}}(k)^\top \mathbb{E}[\mathcal{W}_k^\top \mathcal{W}_k] \tilde{\mathbf{w}}(k) \leq \lambda \| \tilde{\mathbf{w}}(k) \|_2^2. \label{eq:consesnus_error_norm_pt1}
\end{align}
Note from~\eqref{eq:beta_cond} that $0<\lambda$. It also follows from $2\beta(1 - \beta) \lambda_{n-1}(\bar{\mathcal{L}}) > 0$ that $\lambda<1$. Next, taking the square of the norm of $\mathbf{g}(k)$ which is defined below~\eqref{eq:consensus_err}, yields
\begin{align}
    \|\mathbf{g}(k)\|_2^2 &\leq 2\alpha^2 n^2 \|S_k \nabla\mathbf{E}(k)\|_2^2 + 4\alpha \|{S'_k} \mathbf{v}(k)\|_2^2 + 2 \beta^2 \|(\mathcal{L}_k \otimes I_{d_w}) \mathbf{e}(k)\|_2^2, \\
    &\leq 2\alpha^2 n^2 \|S_k\|_2^2 \|\nabla\mathbf{E}(k)\|_2^2 + 4\alpha \|{S'_k}\|_2^2 \| \mathbf{v}(k)\|_2^2 + 2 \beta^2 \|(\mathcal{L}_k \otimes I_{d_w})\|_2^2 \| \mathbf{e}(k)\|_2^2, \\
    &\leq \frac{2\alpha^2 n^2}{4p_m^2} \|\nabla\mathbf{E}(k)\|_2^2 + 4\alpha \| \mathbf{v}(k)\|_2^2 + 2 \beta^2 \| \mathbf{e}(k)\|_2^2, \label{eq:g_exp}
\end{align}
where we used the relations $\|S_k\|_2^2\leq \frac{1}{2p_m}$, $\|S_k\|_2^2 = \|\mathcal{L}_k\|_2^2 = 1$. Further, from Assumption~\ref{assump:bounded_grad} and~\eqref{eq:ET_bound}, it follows that $\mathbb{E}[\|\nabla \mathbf{E}(k)\|_2^2] \leq n\mu_g$ and $\mathbb{E}[\|\mathbf{e}(k) \|_2^2] \leq \frac{n\mu_e}{(k+1)^{\delta_e }}$. Additionally, it can be shown that $\|\mathbf{v}(k)\| \leq n^2d_w$ (refer to (S84) in~\cite{parayil2020decentralized}). Thereafter, taking the expectation of~\eqref{eq:g_exp} and substituting the above bounds results in the following expression.
\begin{align}
    &\mathbb{E}[\| \mathbf{g}{(k)}\|_2^2|\mathcal{F}_k] \leq 2\left( \frac{\alpha^2n^2\mu_g}{4p_m^2} + 2\alpha n^2d_w \right) + \frac{2{\beta}^2n\mu_e}{(k+1)^{\delta_e}}. \label{eq:consesnus_error_norm_pt2}
\end{align}

Now, recall the identity $(x+y)^2\leq (\theta+1)x^2 + \left(\frac{\theta+1}{\theta} \right) y^2$ for any $x,y,\theta \in \mathbb{R}$ and $\theta>0$. We use this identity with $\theta=\sqrt{\lambda}^{-1}-1>0$ on~\eqref{eq:consensus_err_norm}, subsequently take the conditional expectation $\mathbb{E}[\cdot|\mathcal{F}_k]$, and substitute~\eqref{eq:consesnus_error_norm_pt1} and~\eqref{eq:consesnus_error_norm_pt2}. Further, taking the total expectation yields
\begin{align}
    &\mathbb{E}[\|\tilde{\mathbf{w}}(k+1)\|_2^2] \leq \sqrt{\lambda} \mathbb{E}[\|\tilde{\mathbf{w}}(k)\|_2^2] + \frac{1}{1-\sqrt{\lambda}} \frac{2\beta^2n\mu_e}{(k+1)^{\delta_e}} + \frac{2\alpha n^2 (\frac{\alpha\mu_g}{4p_m^2} + 2d_w)}{1-\sqrt{\lambda}}. \label{eq:consesnus_error_norm_it}
\end{align}
Finally, applying Lemma~\ref{lemma:1} in Appendix to~\eqref{eq:consesnus_error_norm_it} results in the  consensus error bound~\eqref{eq:consensus_error_bound}.
\end{proof}

\subsection{Convergence analysis} \label{sec:Result_convergence}
We denote by $p(\bar{\bm{w}}(k))$ the probability distribution of $\bar{\bm{w}}(k)$ 
admitted by the average dynamics~\eqref{eq:average_dynamics} and analyze its  evolution.
To do so, we first reformulate~\eqref{eq:average_dynamics} as a stochastic differential equation (SDE). For any $t\in[t_k,t_{k+1})$ where $t_k = k\alpha$, the SDE form of~\eqref{eq:average_dynamics} is given by
\begin{align}
\begin{split}
    &d\bar{\bm{w}}(t) = - \Big( \nabla E(\bar{\bm{w}}(t_k)) - \xi( \bar{\bm{w}}(t_k), \mathcal{A}_k) + \zeta( \bar{\bm{w}}(t_k), \tilde{\mathbf{w}}(t_k), \mathcal{A}_k) \Big) dt + \sqrt{2} d\bm{B}{(t)}, \label{eq:avg_dyn_SDE}
\end{split}
\end{align}
where $\bm{B}{(t)}$ represents a $d_w$ dimensional Brownian motion, $\bar{\bm{w}}(t_k) = \bar{\bm{w}}(k)$, and $\tilde{\mathbf{w}}(t_k) = \tilde{\mathbf{w}}(k)$. 
{Denote by $p_t(\bar{\bm{w}})$ the distribution of $\bar{\bm{w}}(t)$ from~\eqref{eq:avg_dyn_SDE}.}
{Since the gradient terms in~\eqref{eq:avg_dyn_SDE} remain constant within $t\in[t_k,t_{k+1})$, $p_{t_{k+1}}(\bar{\bm{w}})$ is the same as $p(\bar{\bm{w}}(k+1))$ from~\eqref{eq:average_dynamics}, $\forall k\geq 0$.  Thus, we analyze the evolution of $p_{t}(\bar{\bm{w}})$ from~\eqref{eq:avg_dyn_SDE}.} Let $y_{k,1} = \bar{\bm{w}}(t_k)$, $y_{k,2} = \tilde{\mathbf{w}}(t_k)$, $y_{k,3} = \mathcal{A}_k$, and $y_k = [y_{k,1}^\top, y_{k,2}^\top, y_{k,3}^\top ]^\top$. Using the Fokker Planck (FP) equation for the SDE in~\eqref{eq:avg_dyn_SDE} we have
\begin{align}
    &\frac{\partial p_t(\bar{\bm{w}}|y_k)}{\partial t} = - \nabla  \cdot \bigg[ p_t(\bar{\bm{w}}|y_k) \Big( -\nabla E(y_{k,1}) + \xi(y_{k,1},y_{k,3}) - \zeta(y_k) \Big) \bigg] + \nabla^2 p_t(\bar{\bm{w}}|y_k). \label{eq:FP_eqn}
\end{align}
Marginalizing out $y_k$ from~\eqref{eq:FP_eqn}, we get the evolution of $p_t(\bar{\bm{w}})$ for $t \in [t_k,t_{k+1})$ corresponding to any $k \geq 0$ as 
\begin{align}
    &\frac{\partial p_t(\bar{\bm{w}})}{\partial t} = - \nabla  \cdot \bigg[ \iint \sum_{y_{k,3}\in\mathbf{A}} p_t(\bar{\bm{w}}|y_k) \Big(-\nabla E(\bar{\bm{w}}(t_k)) + \xi(y_{k,1},y_{k,3}) - \zeta(y_k) \Big) p(y_k) dy_{k,1} \, dy_{k,2}\bigg] + \nabla^2 p_t(\bar{\bm{w}}), \label{eq:FP_eqn_mar}
\end{align}
where $\mathbf{A}$ is the finite set of all possible values of $y_{k,3} = \mathcal{A}_k$, i.e., the set of all possible gossiping partners at any time instant of the universal clock. We next employ the KL divergence between the probability distribution $p_t(\bar{\bm{w}})$ and the target distribution $p^*(\bar{\bm{w}})$, denoted by $F(p_t(\bar{\bm{w}}))$, to prove convergence of the posterior of $\bar{\bm{w}}$ 
in~\eqref{eq:average_dynamics}. Specifically,  $F(p_t(\bar{\bm{w}}))$
 is defined as \begin{align}
    &F(p_t(\bar{\bm{w}})) = \int p_t(\bar{\bm{w}}) \log \left( \frac{p_t(\bar{\bm{w}})}{p^*(\bar{\bm{w}})}\right) d \bar{\bm{w}}. \label{eq:KL_div_def}
\end{align}
Theorem~\ref{thm:convergence} below  establishes that  $F(p_t(\bar{\bm{w}}))$ decreases at the rate of $\mathcal{O}\left( \frac{1}{k^{\delta_e}}\right)$  to a bias $B$ given in~\eqref{eq:B_def}. The proof makes use of~\eqref{eq:FP_eqn_mar} and the LSI~\eqref{eq:LSI_original} to obtain $\dot F(p_t(\bar{\bm{w}}))$ and subsequently bound $F(p_t(\bar{\bm{w}}))$.

\begin{theorem}\label{thm:convergence}
Suppose that Assumptions~\ref{assump:Lips_gradient}--\ref{assump:bounded_moment_sto} hold and that $\alpha$ and $\beta$ satisfy Conditions~\ref{cond:alpha_cond} and~\ref{cond:beta_cond}, respectively., then 
\begin{align}
    &(1) \quad \alpha\rho_U + \ln\sqrt{\lambda} < 0: \quad F(p_{t_{k+1}}(\bar{\bm{w}})) \leq \exp \big(-\alpha\rho_U(k+1)) \big) F(p_{t_0}(\bar{\bm{w}})) + \bar{Y}_1^{'} \exp(-\alpha \rho_U k) + \frac{\bar{Y}_2^{'}}{(k+1)^{\delta_e}} + B,
    \label{eq:Conv_case1} \\
    &(2) \quad \alpha\rho_U + \ln\sqrt{\lambda} > 0: \quad F(p_{t_{k+1}}(\bar{\bm{w}})) \leq \exp \big(-\alpha\rho_U(k+1)) \big) F(p_{t_0}(\bar{\bm{w}})) + \bar{Y}_1^{''} \sqrt{\lambda}^{k+1} + \frac{\bar{Y}_2^{'}}{(k+1)^{\delta_e}} + B,
    \label{eq:Conv_case2}
\end{align}
where $\bar{Y}_1^{'}$, $\bar{Y}_1^{''}$, $\bar{Y}_2^{'}$ and $B$ are positive constants given by
\begin{align}
\begin{split}
    \bar{Y}_1^{'} = \left(1-\frac{1}{\alpha\rho_U + \ln\sqrt{\lambda} }\right) \left(\frac{3\alpha^3L^2\bar{L}^2} {2p_m} + \frac{\alpha L^2}{p_m} \right) Y_1, \label{eq:Y_1_dash_def}
\end{split} \\
\begin{split}
    \bar{Y}_1^{''} = \frac{\sqrt{\lambda}^{k+1}}{\alpha\rho_U + \ln\sqrt{\lambda}} \left(\frac{3\alpha^3L^2\bar{L}^2} {2p_m} + \frac{\alpha L^2}{p_m} \right) Y_1, \label{eq:Y_1_ddash_def}
\end{split} \\
\begin{split}
    \bar{Y}_2^{'} = \left( \alpha\rho_U - \frac{\delta_e}{\bar{k}_2} \right)^{-1} \left(\frac{3\alpha^3L^2\bar{L}^2} {2p_m} + \frac{\alpha L^2}{p_m} \right) Y_2, \label{eq:Y_2_dash_def}
\end{split} \\
\begin{split}
    &B = \frac{\nu}{1-\exp(-\alpha\rho_U)}, \label{eq:B_def} 
\end{split}
\end{align}
in which $\nu=3\alpha^3 \bar{L}^2(C_{\xi} + \bar{L}^2 C_{_{\bar{\bm{w}}}}) + 2\alpha(C_{\xi} + \alpha\bar{L}^2d_w) \left(\frac{3\alpha^3L^2\bar{L}^2} {2p_m} + \frac{\alpha L^2}{p_m} \right) Y_3$ and $Y_3$ is given in~\eqref{eq:O_def}. 
\end{theorem}
\begin{proof}
From~\eqref{eq:KL_div_def} the evolution of $F(p_t(\bar{\bm{w}}))$ is related to $\frac{\partial p_t(\bar{\bm{w}}) }{\partial t}$ by 
\begin{align}
    \dot{F}(p_t(\bar{\bm{w}})) = \bigintssss \left(1+ \nabla \log\left( \frac{p_t(\bar{\bm{w}})}{p^*(\bar{\bm{w}})} \right) \right)\frac{\partial p_t(\bar{\bm{w}})}{\partial t} d \bar{\bm{w}}. \label{eq:KL_div_der}
\end{align}
Substituting~\eqref{eq:FP_eqn_mar} into~\eqref{eq:KL_div_der} and performing all the appropriate marginalization yield
\begin{align}
\begin{split}
    \dot{F}(p_t(\bar{\bm{w}})) &\leq  - \frac{1}{2} \mathbb{E} \left[ \left\|\nabla \log\left( \frac{p_t(\bar{\bm{w}})}{p^*(\bar{\bm{w}})}\right) \right\|_2^2 \right] + 3\alpha^2 \bar{L}^2(C_{\xi} + \bar{L}^2 C_{_{\bar{\bm{w}}}}) + 2(C_{\xi} + \alpha\bar{L}^2d_w) \\
    &\qquad \qquad \qquad + \left(\frac{3\alpha^2L^2\bar{L}^2} {2p_m} + \frac{L^2}{p_m} \right) \mathbb{E}_{p(\tilde{\mathbf{w}}(t_k))} [\|\tilde{\mathbf{w}}(t_k)\|_2^2], \label{eq:KL_der_main}
\end{split}
\end{align}
where $\mathbb{E}_{p_t(\bar{\bm{w}})} [\|\bar{\bm{w}}(t_k)\|_2^2] \leq C_{_{\bar{\bm{w}}}}$ (note that this bound has been proven in Lemma~\ref{lemma:4} in Appendix). For details of the derivation of~\eqref{eq:KL_der_main}, refer to Lemma~\ref{lemma:3} in Appendix. Thereafter, we employ the LSI~\eqref{eq:LSI_original} with $g({\bar{\bm{w}}}) = \displaystyle \frac{p_t({\bar{\bm{w}}})}{p^*({\bar{\bm{w}}})}$ to obtain
\begin{align}
\begin{split}
    F(p_t(\bar{\bm{w}}))=\mathbb{E}_{p_{t}(\bar{\bm{w}})}  \left[  \log\left(\frac{p_t(\bar{\bm{w}})}{p^*(\bar{\bm{w}})}\right) \right] \leq \frac{1}{2 \rho_U} \mathbb{E}_{p_{t}(\bar{\bm{w}})} \left[ \left\|\nabla \log\left(\frac{p_t(\bar{\bm{w}})}{p^*(\bar{\bm{w}})}\right)\right\|_2^2 \right], \label{eq:KL_der_2_main}
\end{split}
\end{align}
which when substituted in~\eqref{eq:KL_der_main} gives a recursive relation in $F(p_t(\bar{\bm{w}}))$ for any $t\in[t_k,t_{k+1})$ as follows:
\begin{align}
\begin{split}
    &\dot{F}(p_t(\bar{\bm{w}})) \leq - \rho_U F(p_t(\bar{\bm{w}})) + 3\alpha^2 \bar{L}^2(C_{\xi} + \bar{L}^2 C_{_{\bar{\bm{w}}}}) + 2(C_{\xi} + \alpha\bar{L}^2d_w) + \left(\frac{3\alpha^2L^2\bar{L}^2} {2p_m} + \frac{L^2}{p_m} \right) \mathbb{E}_{p(\tilde{\mathbf{w}}(t_k))} [\|\tilde{\mathbf{w}}(t_k)\|_2^2]. \label{eq:KL_der_3_main}
\end{split}
\end{align}
Integrating~\eqref{eq:KL_der_3_main} from $t \in (t_k,t_{k+1}]$ and noting that $t_{k+1}-t_k = \alpha$ for any $k\geq 0$ together with the relation $\frac{1-\exp(-\rho_U\alpha)}{\rho_U} \leq \alpha$, we get
\begin{align}
\begin{split}
    F(p_{t_{k+1}}(\bar{\bm{w}})) &\leq \exp(-\alpha \rho_U) F(p_{t_k}(\bar{\bm{w}})) + 3\alpha^3 \bar{L}^2(C_{\xi} + \bar{L}^2 C_{_{\bar{\bm{w}}}}) + 2\alpha(C_{\xi} + \alpha\bar{L}^2d_w) \\
    &\qquad \qquad + \left(\frac{3\alpha^3L^2\bar{L}^2} {2p_m} + \frac{\alpha L^2}{p_m} \right) \mathbb{E}_{p(\tilde{\mathbf{w}}(t_k))} [\|\tilde{\mathbf{w}}(t_k)\|_2^2], \label{eq:KL_dis_evo_1}
\end{split}
\end{align}
Next, substituting~\eqref{eq:consensus_error_bound} in~\eqref{eq:KL_dis_evo_1} yields 
\begin{align}
    F(p_{t_{k+1}}(\bar{\bm{w}})) &\leq \exp(-\alpha\rho_U) F(p_{t_k}(\bar{\bm{w}}))  + \left(\frac{3\alpha^3L^2\bar{L}^2} {2p_m} + \frac{\alpha L^2}{p_m} \right) \left( Y_1 \sqrt{\lambda}^{k+1} + \frac{Y_2}{(k+1)^{\delta_e}} \right) + \nu, \label{eq:KL_dis_evo_2}
\end{align} 
where $\nu=3\alpha^3 \bar{L}^2(C_{\xi} + \bar{L}^2 C_{_{\bar{\bm{w}}}}) + 2\alpha(C_{\xi} + \alpha\bar{L}^2d_w) \left(\frac{3\alpha^3L^2\bar{L}^2} {2p_m} + \frac{\alpha L^2}{p_m} \right) Y_3$. Using~\eqref{eq:KL_dis_evo_2} iteratively yields the following relation.
\begin{align}
\begin{split}
    F(p_{t_{k+1}}(\bar{\bm{w}})) &\leq \exp\left(-\alpha\rho_U(k+1)\right) F(p_{t_0}(\bar{\bm{w}})) + \left(\frac{3\alpha^3L^2\bar{L}^2} {2p_m} + \frac{\alpha L^2}{p_m} \right) \sum_{\ell=0}^{k} \left( Y_1\sqrt{\lambda}^{\ell+1} + \frac{Y_2}{(\ell+1 )^{\delta_e}} \right) \exp \Big(-\alpha\rho_U(k-\ell) \Big) \\
    &\quad + \nu \sum_{\ell=0}^{k} \exp(-\alpha\rho_U\ell). \label{eq:KL_dis_evo_3}
\end{split}
\end{align}
Conducting further analysis for the bounds of the summations in~\eqref{eq:KL_dis_evo_3}, we obtain the convergence rate for~\eqref{eq:average_dynamics} in two cases depending on the sign of $\alpha\rho_U + \ln \sqrt{\lambda}$ (note that $\ln\sqrt{\lambda}<0$ since $\lambda<1$), which are shown in~\eqref{eq:Conv_case1} and~\eqref{eq:Conv_case2}, respectively.
\end{proof}

%% file: Discussion.tex
\section{Discussions} \label{sec:Discussions}

In this section, we highlight some key aspects and insights in our results. Firstly, from~\eqref{eq:consensus_error_bound} we get the rate of consensus as $\mathcal{O}\left(\frac{1}{k^{\delta_e}}\right)$. However,~\eqref{eq:consensus_error_bound} also shows a constant offset $Y_3$ given by~\eqref{eq:O_def} in the asymptotic consensus error. This results from the usage of a constant gradient step size $\alpha$. To keep $Y_3$ low,  we may choose the step size $\alpha$ to be scaled as $\alpha \propto \frac{1}{n^2 d_w}$. Since $Y_3 \sim \mathcal{O}(\alpha)$, using a decreasing step size would lead to decay of this term and asymptotic consensus. We leave the analysis of decreasing step sizes as future work.


Secondly, we conclude from~\eqref{eq:Conv_case1} and~\eqref{eq:Conv_case2} that in either case the rate of convergence is $\mathcal{O}\left( \frac{1}{k^{\delta_e}} \right)$ as well. It is tempting to conclude that  a high value of $\delta_e$ is preferable since it fosters both consensus and convergence rate. However, a high $\delta_e$ value results in a quicker decay of the error threshold in~\eqref{eq:ET_bound}, leading to increased communication overhead as $k$ increases. Thus $\delta_e$ is an important hyperpaprameter trading off the rate of convergence against the communication overhead. 

We also observe from either~\eqref{eq:Conv_case1} or~\eqref{eq:Conv_case2} that in the KL divergence bound, there is a constant bias $B$. From~\eqref{eq:B_def}, the most obvious dependence of $B$ is on the step size $\alpha$. For a sufficiently small $\alpha$, $1-\exp(-\alpha\rho_U) \approx \alpha\rho_U$. Since the least power of $\alpha$ in any of the  terms in $\nu$ is $2$, we have $\nu = \alpha^2 \bar{\nu}$. Hence, $B \approx \frac{\alpha \bar{\nu}}{\rho_U}$. Thus, lowering $\alpha$ is likely to reduce $B$, however, it may also compromise the rate of convergence. Furthermore, $B$ linearly decreases with the reduction in $C_{_{\xi}}$ (variance of the stochasticity of gossip), $C_{_{\bar{\bm{w}}}}$ (variance of the average of samples) and $d_w$ (dimension of the samples). In addition, $B \propto \frac{n^2}{p_m}$, implying that reducing $n$ (the number of agents) and increasing $p_m$ {(the least probability of any agent being active)} reduces the bias. This is intuitive as reducing $n$ or increasing 
$p_m$ lowers the uncertainty in the random selection of gossiping agents which translates to a lower bias.

Finally, note that the last term of $\nu$ (given below~\eqref{eq:B_def}) contains the consensus error offset $Y_3$ while the other terms of $\nu$ are due to the variance from different sources (injected noise, gossip stochasticity, and average of samples).

%% file: Num_Exp.tex
\section{Numerical Experiments}
\label{sec:Num_Exp}

\subsection{Gaussian mixture} \label{sec:GM}
We consider parameter inference of a  Gaussian mixture with tied means \cite{welling2011bayesian}. The Gaussian mixture is given by 
\begin{align} 
    \theta_1 \sim \mathcal{N}(0,\sigma_1^2) \hspace{0.5cm} 
    ; \hspace{0.5cm} 
    \theta_2 \sim \mathcal{N}(0,\sigma_2^2) \label{eq:GM_def1} \\
    x_i \sim \frac{1}{2} \mathcal{N}(\theta_1,\sigma_x^2) + \frac{1}{2} \mathcal{N}(\theta_1 + \theta_2,\sigma_x^2), \label{eq:GM_def2}
\end{align}
where $\sigma_1^2 = 10$, $\sigma_2^2 = 1$, $\sigma_x^2 = 2$ and $\bm{w} \triangleq [\theta_1,\theta_2]^\top \in \mathbb{R}^2$. We draw $100$ data samples $x_i$ from the model with $\theta_1 = 0$ and $\theta_2 = 1$. These data points were equally distributed among $5$ agents, each randomly receiving a set of $20$ data points. The communication topology between the agents is a ring graph. 

\begin{figure}[htpb]
    \centering
    \includegraphics[width=0.8\linewidth]{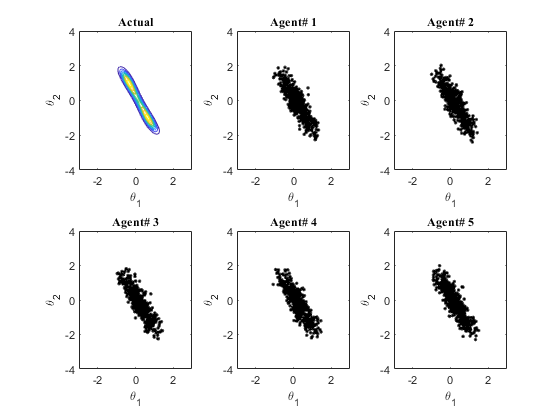}
    \caption{Comparison of the posteriors constructed by the $5$ agents with the actual approximate posterior. The Wassersterin distances between the agents' posteriors with the approximate true posterior are: $0.1089$, $0.0942$, $0.0964$, $0.0963$, and    $0.1073$, respectively.}
    \label{fig:Fig1}
\end{figure}

Simulation results with $1$ Monte Carlo chain for $(100000 \times n)$ iterations is presented. The hyperparameters for the algorithm used are: $\alpha = 1\times10^{-4}$, $\beta = 0.1$, $\mu_e = 8$ and $\delta_e = 0.51$. The samples from the gossip event-triggered algorithm~\eqref{eq:DULA_gossip_ET} 
are compared with an approximated true posterior distribution in Figure~\ref{fig:Fig1}. To compare the accuracy of our results, we used \cite{ECDFdistmeasurealgo} to compute Wasserstein distances as a metric. The presented values of Wasserstein distances are approximations since the target posterior itself is approximated. 

Additional information about the average frequency of gossiping and event-triggering for each agent is listed in Table~\ref{tab:Table1}. Our simulation results suggest that an average (over all agents) of $60\%$ reduction in activity is achieved due to the gossiping protocol, while a significant reduction of more than $80\%$ in communication  is achieved from event-triggering. Note that the percentage reduction in communication due to event-triggering is computed based on the number of times each agent has been active.
\begin{table}[htpb]
\begin{center}
    \begin{tabular}{| c | c | c | c | c | c |  }
\hline
$Agent\#$ & $1$ & $2$ & $3$ & $4$ & $5$    \\ 
\hline \hline
$gos$ & 199481 &  200460  & 200817  &  199912 &  199328  \\
\hline 
$\%\,gos$ & 39.9\% & 40.1\% & 40.2\% & 40.0\% & 39.9\% \\ 
\hline 
$ET$ & 33735 & 33646 & 33695 & 33457 & 33477 \\
\hline
$\%\,ET$ & 16.9\% & 16.8\% & 16.8\% & 16.7\% & 16.8\% \\ 
\hline
\end{tabular}
    \caption{Details about the frequency of gossiping and event-triggering averaged over all $5$ trials for all agents. Here, $gos$ denotes the number of times the agents have been active among the total $500000$ iterations, while $\%\,gos$ denotes gossip as a fraction of the total iterations; $ET$ denotes the number of times the agents have exchanged their samples, while $\%\,ET$ denotes this as a fraction of the total number of times each agent had been active. }
    \label{tab:Table1}
\end{center}
\end{table}

\subsection{Bayesian logistic regression} \label{sec:LR}
We consider the Bayesian logistic regression problem on the UCI ML MAGIC Gamma Telescope dataset\footnote{https://archive.ics.uci.edu/ml/datasets/magic+gamma+telescope}. The dataset contains $19020$ samples with dimension $10$ and each sample describes the registration of high energy gamma particles in a ground-based atmospheric Cherenkov gamma telescope. Each sample has a binary label which signifies either gamma rays or hadron rays. The task is to identify the presence of gamma radiation. 

The entire dataset was split $90$\% into training data and the remaining $10$\% into test data. $10$ chains of Monte Carlo are used. Thereafter, we perform a heterogeneous split where each agent receive different number of data samples with a varying proportions of each category. 
Note, however, that for evaluation of performance by each agent, the accuracy was tested on the same test dataset. Hyperparameters used in this experiments are as follows: $\alpha=10^{-5}$, $\beta=0.5$, $\mu_e=2$ and $\delta_e=0.46$.

\begin{figure}[htpb]
    \centering
    \includegraphics[width=0.8\linewidth]{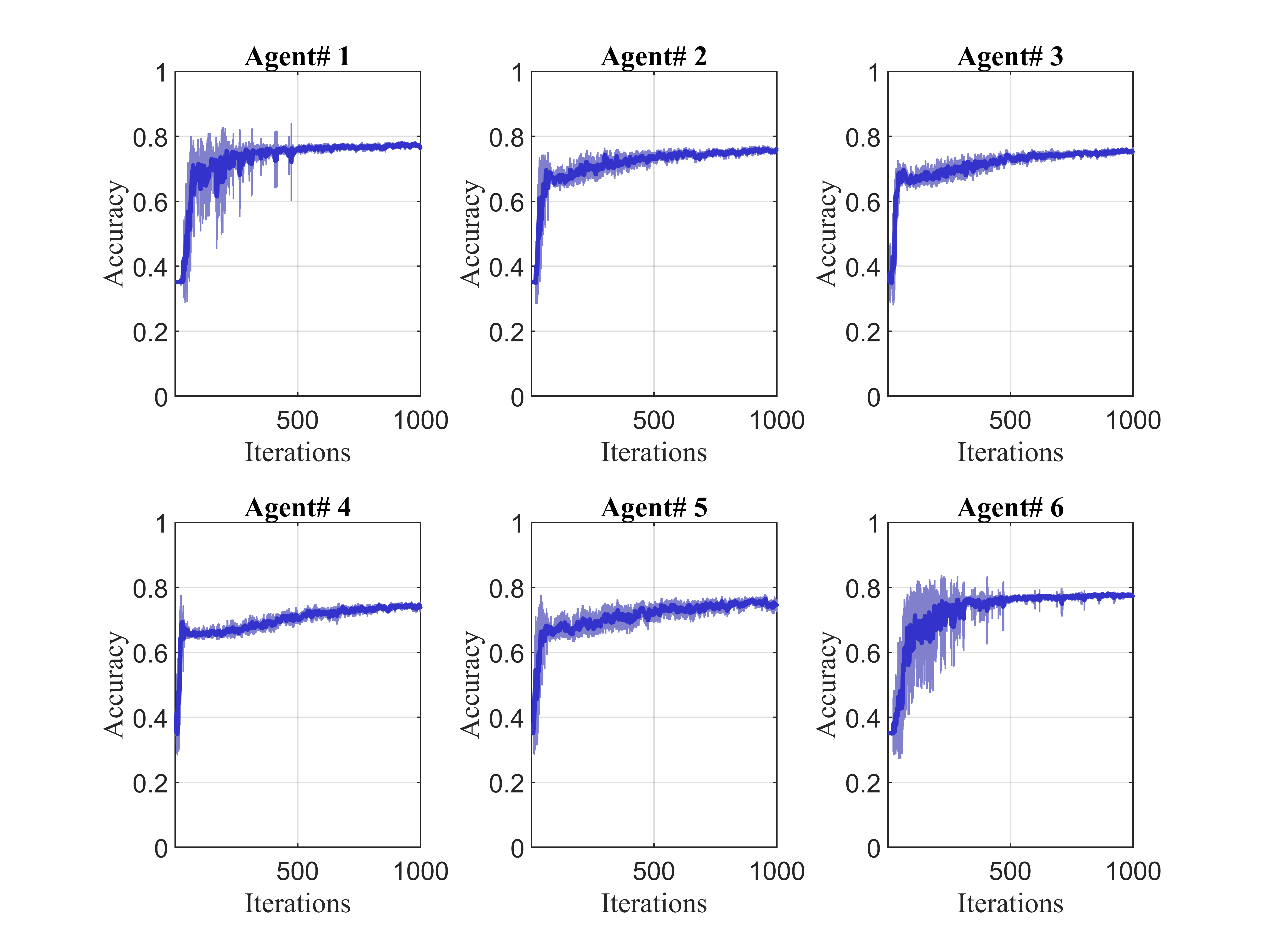}
    \caption{Accuracy results by the $6$ agents for ring graph on the test dataset. Here, ``Iterations'' on the horizontal axis denote the ticks of the universal clock.}
    \label{fig:Fig2}
\end{figure}

Figure~\ref{fig:Fig2} shows that the test data accuracy results for the ring graph for all the agents. Table~\ref{tab:Table2} gives details about the number of times the agents have been active and triggered out of the $1000$ ticks of the universal clock for the ring graph. It shows that gossip reduces activity of agents by roughly $66\%$ and event-triggering reduces the need for communication by another $65\%$ on average. 

\begin{figure}[htpb]
    \centering
    \includegraphics[width=0.6\linewidth]{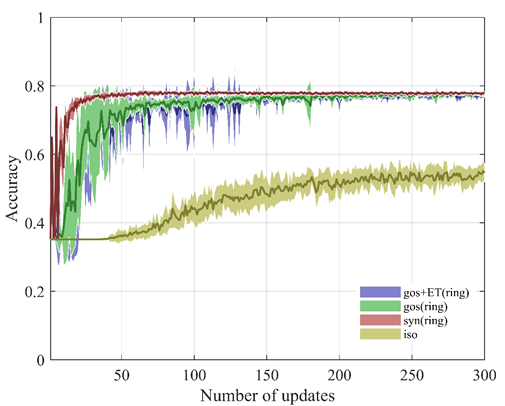}
    \vspace{0.1cm}
    \caption{Accuracy results of Agent$\#\,6$ on the test dataset under different cases. Here,``gos'': gossip; ``ET'': event-triggering; ``syn'': synchronous updates; ``iso'': isolated training; ``(ring)'': ring graph. For a fair comparison between the asynchronous and synchronous updates, we plot the performance with respect to the ``number of updates`` made by Agent$\#\,6$ on the horizontal axis instead of ticks of the universal clock.}
    \label{fig:Fig3}
\end{figure}

{In Figure~\ref{fig:Fig3}, we compare the test accuracy of Agent$\#\,6$ between distributed synchronous, distributed asynchronous, and isolated training. We observe that the distributed synchronous training produces quicker convergence with less variance than the asynchronous algorithms and the isolated training. This is expected as each agent maintains communication with all of its neighbors at each update. Note that the final performance of the synchronous and the asynchronous training is comparable and the net accuracy reached ($\sim 80\%$) is the same. We also observe that there is negligible loss of performance by implementing event-triggering with gossip compared to only gossip. Finally, we clearly observe that distributed training (under all cases) significantly outperforms ($\sim 80\%$ accuracy) isolated training ($\sim 60\%$ accuracy). }

\begin{table}[htpb]
\begin{center}
    \begin{tabular}{| c | c | c | c | c | c | c |  }
\hline
$Agent\#$ & $1$ & $2$ & $3$ & $4$ & $5$ & $6$   \\ 
\hline \hline
$gos$ & 332 &  336  & 335 &  324 &  332 & 339 \\
\hline 
$\%\,gos$ & 33.2\% & 33.6\% & 33.5\% & 32.4\% & 33.2\% & 33.9\% \\ 
\hline 
$ET$ & 117 & 118 & 113 & 109 & 128 & 123 \\
\hline
$\%\,ET$ & 35.3\% & 34.6\% & 34.0\% & 33.6\% & 38.0\% & 36.6\% \\ 
\hline
\end{tabular}
\vspace{0.2cm}
    \caption{Details about the frequency of gossiping and event-triggering averaged over all $10$ trials for all agents for the \emph{ring} graph. Here, $gos$ denotes the number of times the agents have been active among the total $1000$ iterations, while $\%\,gos$ denotes this as a fraction of the total iterations; $ET$ denotes the number of times the agents have broadcasted their samples, while $\%\,ET$ denotes this as a fraction of the total number of times they had been active. }
    \label{tab:Table2}
\end{center}
\end{table}

%% file: Conclusion.tex
\section{Conclusions and future work} \label{sec:Conclusion}

In this paper, we propose an asynchronous distributed Bayesian algorithm for inference over a graph via an event-triggered gossip communication based on the ULA. We derive rigorous convergence guarantees for the proposed algorithm and illustrate its effectiveness using two numerical experiments. Though we obtain good empirical results, our mathematical analysis shows some asymptotic bias in the convergence which stems from the use of a constant step size. Our future work involves the analysis of gossip algorithms with diminishing step sizes and other asynchronous algorithms for distributed Bayesian learning. 

%% file: Appendix.tex
\section*{Appendix} \label{sec:Appendix}

\begin{lemma} \label{lemma:1}
Let $\{y_k\}$ be a non-negative sequence for all $k \in \mathbb{N}$ satisfying
\begin{align}
    y_{k+1} &\leq \sigma y_k + \frac{\mu_{\xi}}{(k+1)^\delta} + c,
\end{align}
where $\sigma, \delta \in (0,1)$ and $\mu_\xi,c>0$. Then the following result holds.
\begin{align}
    y_{k+1} &\leq W_1 \sigma^{k+1} + \frac{W_2}{(k+1)^\delta} + W_3, \label{eq:lemma_cond}
\end{align}
where 
\begin{align}
    W_1 &= y_0 + \mu_\xi \sum_{t=0}^{\bar{t}- 1} \frac{\sigma^{-(t+1)}}{(t+1)^\delta} >0, \label{eq:W1_def} \\
    W_2 &= \sigma^{-1}\mu_\xi \left(|\ln{\sigma}| - \frac{\delta}{(\bar{t}+1)} \right) >0, \label{eq:W2_def} \\
    W_3 &= \frac{c}{1-\delta} >0. \label{eq:W3_def}
\end{align}
\end{lemma}
\begin{proof}
Using~\eqref{eq:lemma_cond} iteratively we obtain the following expression.
\begin{align}
    y_{k+1} &\leq \sigma^{k+1} y_0 + \sum_{t=0}^{k} \left(\frac{\mu_\xi}{ (t+1)^\delta} + c \right) \sigma^{k-t}. \label{eq:lemma_cond_2}
\end{align}
The following analysis is thereafter performed.
\begin{align}
    y_{k+1} &\leq \sigma^{k+1} y_0 + \mu_\xi \sum_{t=0}^{k} \frac{\sigma^{k-t}}{ (t+1)^\delta} + c \sum_{t=0}^{k} \sigma^{k-t}, \\
    &\leq \left(y_0 + \mu_\xi \sum_{t=0}^{\bar{t}-1} \frac{\sigma^{-(t+1) }}{(t+1)^\delta}\right) \sigma^{k+1} + \mu_\xi \sum_{t=\bar{t}}^{k} \frac{\sigma^{k-t}}{(t+1)^\delta} + c \sum_{t=0}^{k} \sigma^t. \label{eq:lemma_cond_3}
\end{align}
The last term of~\eqref{eq:lemma_cond_2} can simply be approximated as 
\begin{align}
    c\sum_{t=0}^{k} \sigma^t \leq c\sum_{t=0}^{\infty} \sigma^t = \frac{c}{1-\sigma}. \label{eq:term_3}
\end{align}
For the second term of~\eqref{eq:lemma_cond_2}, we first choose $\bar{t} = \max\left\{0,\ceil[\Big]{\frac{\delta}{\ln{\sigma}}-1} \right\}$ such that $\frac{\sigma^{k-t}}{(t+1)^\delta}$ is an increasing function for $t \geq \bar{t}$. Thereafter, we use the following approximation.
\begin{align}
    \mu_\xi \sum_{t=\bar{t}}^{k} \frac{\sigma^{k-t}}{(t+1)^\delta} &\leq \mu_\xi \int_{\bar{t}}^{k+1} \frac{\sigma^{k-t}}{(t+1)^\delta} dt \leq \frac{\mu_\xi \sigma^{-1} \left(|\ln{\sigma}| - \frac{\delta}{(\bar{t}+1)} \right)^{-1}}{(k+1)^\delta}. \label{eq:term_2}
\end{align}
In order to perform the last approximation in~\eqref{eq:term_2}, we first observe 
\begin{align}
    \frac{d}{dt} \left(\frac{\sigma^{k-t}} {(t+1)^\delta} \right) &= \frac{\sigma^{k-t}}{(t+1)^ \delta} \left(|\ln{\sigma}| - \frac{\delta}{t+1} \right) \geq \frac{\sigma^{k-t}}{(t+1)^ \delta} \left(|\ln{\sigma}| - \frac{\delta}{\bar{t}+1} \right), \quad \forall \, t\geq \bar{t}.
\end{align}
Note that in the interval $t \geq \bar{t}$, $\frac{\sigma^{k-t}}{(t+1)^ \delta}$ is increasing, hence $|\ln{\sigma}| - \frac{\delta}{t+1}$ is positive. Thus, 
\begin{align}
    \int_{\bar{t}}^{k+1}  \frac{\sigma^{k-t}}{(t+1)^ \delta} &\leq \left( |\ln{\sigma}| - \frac{\delta}{\bar{t}+1} \right)^{-1} \left(\frac{\sigma^{k-t}}{(t+1)^ \delta} \right) \bigg|_{\bar{t}}^{k+1} \leq \left( |\ln{\sigma}| - \frac{\delta}{\bar{t}+1} \right)^{-1} \frac{\sigma^{-1}}{(k+1)^ \delta}.
\end{align}
Next, substituting~\eqref{eq:term_3} and~\eqref{eq:term_2} in~\eqref{eq:lemma_cond_3} and using the definitions of $W_1$-$W_3$ from~\eqref{eq:W1_def}-\eqref{eq:W3_def} respectively yields the result in~\eqref{eq:lemma_cond}.
\end{proof}

\begin{lemma} \label{lemma:3}
Given Assumptions~\ref{assump:Lips_gradient} and~\ref{Assump:LSI}, we have 
\begin{align}
\begin{split}
    \dot{F}(p_t(\bar{\bm{w}})) &\leq - \frac{1}{2} \mathbb{E} \left[ \left\|\nabla \log\left( \frac{p_t(\bar{\bm{w}})}{p^*(\bar{\bm{w}})}\right) \right\|_2^2 \right] + 3\alpha^2 \bar{L}^2(C_{\xi} + \bar{L}^2 C_{_{\bar{\bm{w}}}}) + 2(C_{\xi} + \alpha\bar{L}^2d_w) \\
    &\qquad \qquad + \left(\frac{3\alpha^2L^2\bar{L}^2} {2p_m} + \frac{L^2}{p_m} \right) \mathbb{E}_{p(\tilde{\mathbf{w}}(t_k))} [\|\tilde{\mathbf{w}}(t_k)\|_2^2]. \label{eq:F_dot_2}
\end{split}
\end{align}
\end{lemma}
\begin{proof}
Following a similar analysis as in Section S4 in~\cite{parayil2020decentralized},~\eqref{eq:FP_eqn_mar} leads to the expression below.
\begin{align}
    \frac{\partial p_t(\bar{\bm{w}})}{\partial t} &= \nabla \cdot \left[p_t(\bar{\bm{w}}) \nabla \log \left( \frac{p_t(\bar{\bm{w}}) }{p^*(\bar{\bm{w}})}\right) \right] - \nabla \cdot \Bigg[\iint \sum_{y_{k,3}\in\mathbf{A}} \Big(\nabla E(\bar{\bm{w}}) - \nabla E(y_{k,1}) + \xi(y_{k,1},y_{k,3}) - \zeta(y_k)\Big) p_t(\bar{\bm{w}},y_k)\times \nonumber \\
    &\qquad \qquad dy_{k,1} dy_{k,2} \Bigg], \\
    &= \nabla \cdot f_t(\bar{\bm{w}}) - \nabla \cdot g_t(\bar{\bm{w}},y_k), \label{eq:FP_eqn_2}
\end{align}
where 
\begin{align}
    f_t(\bar{\bm{w}}) &= \left[p_t(\bar{\bm{w}}) \nabla \log \left( \frac{p_t(\bar{\bm{w}}) }{p^*(\bar{\bm{w}})}\right) \right] \\
    g_t(\bar{\bm{w}},y_k) &= \Big[\iint \sum_{y_{k,3}\in\mathbf{A}} \big(\nabla E(\bar{\bm{w}}) - \nabla E(y_{k,1}) + \xi(y_{k,1},y_{k,3}) - \zeta(y_k)\big) p_t(\bar{\bm{w}},y_k) dy_{k,1} dy_{k,2} \Big]
\end{align}
Thereafter, substituting~\eqref{eq:FP_eqn_2} into~\eqref{eq:KL_div_der} and making use of Lemma S5 from~\cite{parayil2020decentralized} yields
\begin{align}
     \dot{F}(p_t(\bar{\bm{w}})) &= - \int \nabla \log\left( \frac{p_t(\bar{\bm{w}})}{p^*(\bar{\bm{w}})}\right)^\top f_t(\bar{\bm{w}}) d\bar{\bm{w}} + \int \nabla \log\left( \frac{p_t(\bar{\bm{w}})}{p^*(\bar{\bm{w}})}\right)^\top g_t(\bar{\bm{w}},y_k) d\bar{\bm{w}}. \label{eq:F_dot}
\end{align}
Again, from (S116) in~\cite{parayil2020decentralized}, the first term in~\eqref{eq:F_dot} can be simplified as below. 
\begin{align}
    \int \nabla \log\left( \frac{p_t(\bar{\bm{w}})}{p^*(\bar{\bm{w}})}\right)^\top f_t(\bar{\bm{w}}) d\bar{\bm{w}} = \mathbb{E} \left[ \left\|\nabla \log\left( \frac{p_t(\bar{\bm{w}})}{p^*(\bar{\bm{w}})}\right) \right\|_2^2 \right]. \label{eq:F_dot_f_term}
\end{align}
Similarly, for the second term in~\eqref{eq:F_dot} we get 
\begin{align}
    &\int \nabla \log\left( \frac{p_t(\bar{\bm{w}})}{p^*(\bar{\bm{w}})}\right)^\top g_t(\bar{\bm{w}},y_k) d\bar{\bm{w}} \leq \frac{1}{2} \mathbb{E} \left[ \left\|\nabla \log\left( \frac{p_t(\bar{\bm{w}})}{p^*(\bar{\bm{w}})}\right) \right\|_2^2 \right] + \bar{L}^2 \iint \|\bar{\bm{w}} - y_{k,1}\|_2^2 p_t(\bar{\bm{w}},y_k) dy_k d\bar{\bm{w}} \nonumber \\
    &\qquad \qquad + 2\iint \sum_{y_{k,3}\in \mathbf{A}} \|\zeta(y_k)\|_2^2 p_t( \bar{\bm{w}},y_k) dy_k d\bar{\bm{w}} + 2\iint \sum_{y_{k,3}\in \mathbf{A}} \|\xi(y_{k,1},y_{k,3})\|_2^2 p_t( \bar{\bm{w}},y_k) dy_k d\bar{\bm{w}}. \label{eq:F_dot_g_term_1}
\end{align}
We next analyse the individual term on the right hand side of~\eqref{eq:F_dot_g_term_1} separately. From Assumption~\ref{assump:bounded_moment_sto},
\begin{align}
    2\iint \sum_{y_{k,3}\in \mathbf{A}} \|\xi(y_{k,1},y_{k,3})\|_2^2 p_t( \bar{\bm{w}},y_k) dy_k d\bar{\bm{w}} &\leq 2C_\xi. \label{eq:F_dot_g_term_2}
\end{align}
Now,  
\begin{align}
    \sum_{y_{k,3} \in \mathbf{A}} \|\zeta(y_k)\|_2^2 p_t(y_{k,3}) &= \sum_{\mathcal{A}_k \in \mathbf{A}} \left\| \sum_{i\in y_{k,3}} \frac{1}{2p_i} \Big(\nabla E_i(\bm{w}_i(k)) - \nabla E_i(\bar{\bm{w}}(k)) \Big)\right\|_2^2 p_t(y_{k,3}) &&\leq \sum_{y_{k,3} \in \mathbf{A}} \left\| \sum_{i\in y_{k,3}} \frac{L_i}{2p_i} \tilde{\bm{w}}_i(k) \right\|_2^2 p_t(y_{k,3}), \\
    &\leq \sum_{y_{k,3} \in \mathbf{A}} \sum_{i\in y_{k,3}} \frac{L_i^2}{2p_i^2}\|\tilde{\bm{w}}_i(k)\|_2^2 p_t(y_{k,3}) &&= \sum_{i=1}^{n} \frac{L_i^2}{2p_i^2}\|\tilde{\bm{w}}_i(k)\|_2^2 p_i, \\
    &\leq \frac{L^2}{2p_m} \sum_{i=1}^{n} \|\tilde{\bm{w}}_i(k)\|_2^2 &&= \frac{L^2}{2p_m}\|\tilde{\mathbf{w}}(k)\|_2^2.
\end{align}
We next provide a brief explanation of the second step of the above inequality. Consider the $i$-th agent, the probability of its link with $j_k$ ($j_k$ is any neighbor of $i_k$) being active is $\frac{1}{n}\left( \frac{1}{|\mathcal{N}_i|} + \frac{1}{|\mathcal{N}_j|}\right)$. Thus, the total probability of links containing $i_k$ being active will be $\sum_{j \in \mathcal{N}_i} \frac{1}{n}\left( \frac{1}{|\mathcal{N}_i|} + \frac{1}{|\mathcal{N}_j|}\right) = \frac{1}{n} \left( 1 + \sum_{j\in \mathcal{N}_i} \frac{1}{|\mathcal{N}_j|}\right) = p_i$. Hence, the coefficient of $\|\tilde{\bm{w}}_i\|_2^2$ in $\displaystyle\sum_{y_{k,3}\in\mathbf{A}} \sum_{i\in y_{k,3}} \frac{L_i^2}{2p_i^2} \| \tilde{\bm{w}}_i\|_2^2 p_t(\mathcal{A}_k)$ is $\frac{L_i^2}{2p_i^2} \times p_i = \frac{L_i^2}{2p_i}$ for any $i$-th agent.
Therefore,
\begin{align}
    2\iint \sum_{y_{k,3}\in \mathbf{A}} \|\zeta(y_k)\|_2^2 p_t( \bar{\bm{w}},y_k) dy_k d\bar{\bm{w}} &\leq 2\iint \sum_{y_{k,3}\in \mathbf{A}} \|\zeta(y_k)\|_2^2 p_t(y_{k,3}) p_t(\bar{\bm{w}},y_{k,1},y_{k,2}|y_{k,3}) dy_{k,1} dy_{k,2} d\bar{\bm{w}}, \\
    &\leq 2\iint \frac{L^2}{2p_m} \| \tilde{\mathbf{w}}(k)\|_2^2 p_t(\bar{\bm{w}},y_{k,1},y_{k,2}) dy_{k,1} dy_{k,2} d\bar{\bm{w}}, \\
    &\leq \frac{L^2}{p_m} \mathbb{E}_{p_t(\tilde{\mathbf{w}}(t_k))}\|\tilde{\mathbf{w}}(t_k)\|_2^2.
    \label{eq:F_dot_g_term_3}
\end{align}
Next, we have
\begin{align}
    \|\bar{\bm{w}} - y_{k,1}\|_2^2 &= \| -\nabla E(y_{k,1})(t-t_k) + \xi(y_{k,1}, y_{k,3}) (t-t_k) - \zeta(y_k)(t-t_k) + \sqrt{2} (\bm{B}(t) - \bm{B}(t_k))\|_2^2, \\
    &\leq (t-t_k)^2 \|-\nabla E(y_{k,1}) + \xi(y_{k,1}, y_{k,3}) - \zeta(y_k)\|_2^2  + 2 \|\bm{B}(t) - \bm{B}(t_k)\|_2^2 \nonumber \\
    &\qquad + 2\sqrt{2} (t-t_k) (\bm{B}(t) - \bm{B}(t_k))^\top (-\nabla E(y_{k,1}) + \xi(y_{k,1}, y_{k,3})- \zeta(y_k)), \\
    &\leq 3\alpha^2 \|\nabla E(y_{k,1}) \|_2^2+ 3\alpha^2 \|\xi(y_{k,1}, y_{k,3})\|_2^2 + 3\alpha^2\| \zeta(y_k)\|_2^2  + 2 \|\bm{B}(t) - \bm{B}(t_k)\|_2^2 \nonumber \\
    &\qquad + 2\sqrt{2} (t-t_k) (\bm{B}(t) - \bm{B}(t_k))^\top (-\nabla E(y_{k,1}) + \xi(y_{k,1}, y_{k,3})- \zeta(y_k)),
\end{align}
where we use $t-t_k \leq \alpha$. Thereafter, we have
\begin{align}
    2\iint \sum_{y_{k,3}\in\mathbf{A}} \| \bm{B}(t) - \bm{B}(t_k)\|_2^2 p_t(\bar{\bm{w}},y_k) d\bar{\bm{w}}dy_k &\leq 2\alpha d_w, \label{eq:C1}
\end{align}
and 
\begin{align}
    2\iint \sum_{y_{k,3}\in\mathbf{A}} 2\sqrt{2} (t-t_k) \Big(\bm{B}(t) - \bm{B}(t_k)\Big)^\top \Big(-\nabla E(y_{k,1}) + \xi(y_{k,1}, y_{k,3})- \zeta(y_k)\Big) p_t(\bar{\bm{w}},y_k) d\bar{\bm{w}}dy_k &= 0, \label{eq:C2}
\end{align}
Refer to (S138) for~\eqref{eq:C1} and (S141) for~\eqref{eq:C2} in~\cite{parayil2020decentralized} for details. Finally,
\begin{align}
    3\alpha^2 \iint \sum_{y_{k,3}\in\mathbf{A}} \|\nabla E(y_{k,1}) \|_2^2 p_t(\bar{\bm{w}},y_k) d\bar{\bm{w}}dy_k &\leq 3\alpha^2 \bar{L}^2 \iint \sum_{y_{k,3}\in\mathbf{A}} \|y_{k,1}\|_2^2 p_t(\bar{\bm{w}},y_k) d\bar{\bm{w}}dy_k \\
    &= 3\alpha^2 \bar{L}^2 \mathbb{E}_{p_t(\bar{\bm{w}}(t_k))} [\|y_{k,1}\|_2^2] \leq 3\alpha^2 \bar{L}^2 C_{_{\bar{\bm{w}}}},
\end{align}
where we assume without loss of generality that $\nabla E(\mathbf{0}) = \mathbf{0}$ and $\|\nabla E(y_{k,1})\| = \|\nabla E(y_{k,1}) - \nabla E(\mathbf{0})\| \leq \bar{L} \|y_{k,1}\|$ and use the bound $\mathbb{E}_{p_t(\bar{\bm{w}}(t_k))} [\|y_{k,1}\|_2^2] = \mathbb{E}_{p_t(\bar{\bm{w}}(t_k))} [\|\bar{\bm{w}}(t_k)\|_2^2] \leq \bar{\bm{w}}(t_k)$ (refer to Lemma~\ref{lemma:4}).
Thus, 
\begin{align}
    &\iint \sum_{y_{k,3}\in\mathbf{A}} \Big(3\alpha^2 \|\nabla E(y_{k,1}) \|_2^2+ 3\alpha^2 \|\xi(y_{k,1}, y_{k,3})\|_2^2 + 3\alpha^2\| \zeta(y_k)\|_2^2 \Big) p_t(\bar{\bm{w}},y_k) d\bar{\bm{w}}dy_k \\
    &\quad \leq 3\alpha^2 \bar{L}^2 C_{_{\bar{\bm{w}}}} + 3\alpha^2C_{\xi} + \frac{3\alpha^2L^2}{2p_m} \mathbb{E}_{p_t(\tilde{\mathbf{w}}(t_k))}\|\tilde{\mathbf{w}}(t_k)\|_2^2. \label{eq:C3}
\end{align}
Therefore, using~\eqref{eq:C1},~\eqref{eq:C2} and~\eqref{eq:C3} yields
\begin{align}
    \bar{L}^2 \iint \|\bar{\bm{w}} - y_{k,1}\|_2^2 p_t(\bar{\bm{w}},y_k) dy_k d\bar{\bm{w}} &\leq 2\alpha\bar{L}^2d_w + 3\alpha^2 \bar{L}^4 C_{_{\bar{\bm{w}}}} + 3\alpha^2 \bar{L}^2 C_{\xi} + \frac{ 3\alpha^2 L^2 \bar{L}^2}{2p_m} \mathbb{E}_{p(\tilde{\mathbf{w}}(t_k))} [\|\tilde{\mathbf{w}}(t_k)\|_2^2]). \label{eq:F_dot_g_term_4}
\end{align}
Now, substituting~\eqref{eq:F_dot_g_term_2}, \eqref{eq:F_dot_g_term_3} and~\eqref{eq:F_dot_g_term_4} in~\eqref{eq:F_dot_g_term_1} results in
\begin{align}
    \int \nabla \log\left( \frac{p_t(\bar{\bm{w}})}{p^*(\bar{\bm{w}})}\right)^\top g_t(\bar{\bm{w}},y_k) d\bar{\bm{w}} &\leq \frac{1}{2} \mathbb{E} \left[ \left\|\nabla \log\left( \frac{p_t(\bar{\bm{w}})}{p^*(\bar{\bm{w}})}\right) \right\|_2^2 \right] + 3\alpha^2 \bar{L}^2(C_{\xi} + \bar{L}^2 C_{_{\bar{\bm{w}}}}) + 2C_{\xi} + 2\alpha\bar{L}^2d_w \nonumber \\
    &\qquad \qquad + \left(\frac{3\alpha^2L^2\bar{L}^2} {2p_m} + \frac{L^2}{p_m} \right) \mathbb{E}_{p(\tilde{\mathbf{w}}(t_k))} [\|\tilde{\mathbf{w}}(t_k)\|_2^2]. \label{eq:F_dot_g_term_5}
\end{align}
Substituting~\eqref{eq:F_dot_f_term} and~\eqref{eq:F_dot_g_term_5} in~\eqref{eq:F_dot} gives~\eqref{eq:F_dot_2}.
\end{proof}

\begin{lemma} \label{lemma:4}
Suppose $\bar{\bm{w}}^* \sim p^*$ satisfy $\mathbb{E}[\|\bar{\bm{w}}^*\|_2^2] \leq c_1 < \infty$ and $F(p_{t_0}(\bar{\bm{w}})) \leq c_2 < \infty$, and Assumptions~\ref{assump:Lips_gradient}-\ref{Assump:LSI} and Conditions~\ref{cond:alpha_cond}-\ref{cond:beta_cond} hold. Then there exists $C_{_{\bar{\bm{w}}}}$ such that 
\begin{align}
    \mathbb{E}_{p_{t_k}(\bar{\bm{w}})} [\|\bar{\bm{w}}(t_k)\|_2^2] \leq C_{_{\bar{\bm{w}}}}, \quad  \forall \, t_k\geq 0.
\end{align}
\end{lemma}
\begin{proof}
We start with assuming $\mathbb{E} [\|\bar{\bm{w}}(t_j)\|_2^2] \leq C_{_{\bar{\bm{w}}}}$ for all $0<t_j<t_k<\infty$ and then make use of induction to prove $\mathbb{E}[\| \bar{\bm{w}}(t_{k+1})\|_2^2] \leq C_{_{\bar{\bm{w}}}}$. 

To that end, we couple $\bar{\bm{w}}^*\sim p^*$ optimally with $\bar{\bm{w}}(t) \sim p_t(\bar{\bm{w}})$, i.e., $(\bar{\bm{w}}(t), \bar{\bm{w}}^*) \sim \Phi \in \tau_{opt} (p_t(\bar{\bm{w}}(t)), p^*)$. Thus,
\begin{align}
\begin{split}
    \mathbb{E}_{p_{t_{k+1}}(\bar{\bm{w}})} [\|\bar{\bm{w}}(t_{k+1})\|_2^2] &= \mathbb{E}_{\Phi} [\|\bar{\bm{w}}^* + \bar{\bm{w}}(t_{k+1}) - \bar{\bm{w}}^* \|_2^2] = 2\mathbb{E}_{p^*} [\|\bar{\bm{w}}^*\|_2^2] + 2\mathbb{E}_{\Phi} [\|\bar{\bm{w}}(t_{k+1}) - \bar{\bm{w}}^*\|_2^2],  \\
    &\leq 2c_1 + 2\mathbb{W}_2^2(p_{t_{k+1}}(\bar{\bm{w}}),p^*) = 2c_1 + \frac{4}{\rho_U} F(p_{t_{k+1}}(\bar{\bm{w}})), \label{eq:w_bar_bound_1}
\end{split}
\end{align}
where $\mathbb{W}_2^2(\cdot\,,\cdot)$ denotes the Wasserstein distance between two distributions. Assuming that $\mathbb{E}_{p_{t}} [\|\bar{\bm{w}}(t)\|_2^2] \leq C_{_{\bar{\bm{w}}}}$ for all $t\leq t_k$ the form of the bounds for $F(p_{t_{k+1}}(\bar{\bm{w}}))$ results in the same expressions given in~\eqref{eq:Conv_case1} and~\eqref{eq:Conv_case2}. Substituting them in~\eqref{eq:w_bar_bound_1} results in
\begin{align}
    \begin{split}
    \mathbb{E}_{p_{t_{k+1}}(\bar{\bm{w}})} [\|\bar{\bm{w}}(t_k+1)\|_2^2] &\leq 2c_1 + \frac{4}{\rho_U} \Bigg(\frac{F(p_{t_0}(\bar{\bm{w}}))}{\exp(\alpha \rho_U(k+1))} + \frac{\bar{Y}_2^{'}}{(k+1)^{\delta_e}} + {\max}\left\{ \frac{\bar{Y}_1^{'}}{\exp(\alpha\rho_U k)} , \bar{Y}_1^{''}\sqrt{\lambda}^{k+1} \right\} \\
    &\qquad \qquad \qquad + \frac{\nu}{1-\exp(-\alpha \rho_U)} \Bigg), 
    \end{split}\\
    &\leq 2c_1 + \frac{4}{\rho_U} \Bigg( F(p_{t_0}(\bar{\bm{w}})) + \bar{Y}^M + \frac{\nu}{1-\exp(-\alpha \rho_U)} \Bigg), \label{eq:w_bar_bound_2} 
\end{align}
where $\bar{Y}^M = \bar{Y}_2^{'} + \max\left\{ \bar{Y}_1^{'}, \bar{Y}_1^{''} \right\}$. To use induction, we need to have $\mathbb{E}_{p_{t_{k+1}}(\bar{\bm{w}})} [\|\bar{\bm{w}}(t_{k+1})\|_2^2] \leq C_{_{\bar{\bm{w}}}}$, which is guaranteed if (after substituting the expression of $\nu$ given below~\eqref{eq:B_def})
\begin{align}
    \bar{C}_{_{\bar{\bm{w}}}} \leq C_{_{\bar{\bm{w}}}}, 
\end{align}
where
\begin{align}
    \bar C_{_{\bar{\bm{w}}}} &= \frac{1}{ 1- \frac{8\alpha^3\bar{L}^4}{ \rho_U(1-\exp(- \alpha \rho_U))}}
    \Bigg[ 2c_1 + \frac{4}{\rho_U} \Bigg( F(p_{t_0}(\bar{\bm{w}}))+\bar{Y}^{M} + \frac{2\alpha^2\bar{L}^2d_w + 4\alpha^3\bar{L}^2C_{_{\xi}} + \left( \frac{\alpha^3\bar{L}^2L^2}{p_m} + \frac{\alpha L^2}{4p_m} \right) Y_3}{1-\exp(-\alpha \rho_U)} \Bigg) \Bigg]. 
\end{align}
Note from~\eqref{eq:alpha_cond} that $1- \frac{8\alpha^3\bar{L}^4}{ \rho_U(1-\exp(- \alpha \rho_U))} >0$.
Finally, we conclude that choosing $C_{_{\bar{\bm{w}}}}$ as
\begin{align}
\begin{split}
    &C_{_{\bar{\bm{w}}}} = \max\left\{ \, \mathbb{E}_{p_{t_0}(\bar{\bm{w}})} [\| \bar{\bm{w}}\|_2^2],~~\bar C_{_{\bar{\bm{w}}}}\right\}
\end{split}
\end{align}
ensures $\mathbb{E}_{p_{t_k}(\bar{\bm{w}})} [\|\bar{\bm{w}}(t_k)\|_2^2] \leq C_{_{\bar{\bm{w}}}}$ for all $t_k \geq 0$. Thus, the second moment of the average sample $\bar{\bm{w}}$ is always bounded.
\end{proof}